# Horizontal Federated Learning and Secure Distributed Training for Recommendation System with Intel SGX

Siyuan Hui
Intel Corporation
Shanghai, China
siyuan.hui@intel.com

Yuqiu Zhang
Intel Corporation
Shanghai, China
yuqiu.zhang@intel.com

Albert Hu
Intel Corporation
Shanghai, China
albert.hu@intel.com

Edmund Song
Intel Corporation
Shanghai, China
edmund.song@intel.com

## ABSTRACT

With the advent of big data era and the development of artificial intelligence and other technologies, data security and privacy protection have become more important. Recommendation systems have many applications in our society, but the model construction of recommendation systems is often inseparable from users' data. Especially for deep learning-based recommendation systems, due to the complexity of the model and the characteristics of deep learning itself, its training process not only requires long training time and abundant computational resources but also needs to use a large amount of user data, which poses a considerable challenge in terms of data security and privacy protection. How to train a distributed recommendation system while ensuring data security has become an urgent problem to be solved.

In this paper, we implement two schemes, Horizontal Federated Learning and Secure Distributed Training, based on Intel SGX(Software Guard Extensions) [1], an implementation of a trusted execution environment [2], and TensorFlow framework [3], to achieve secure, distributed recommendation system-based learning schemes in different scenarios. We experiment on the classical Deep Learning Recommendation Model (DLRM), which is a neural network-based machine learning model designed for personalization and recommendation [4], and the results show that our implementation introduces approximately no loss in model performance. The training speed is within acceptable limits.

## 1 INTRODUCTION

The goal of the recommendation system is to infer users' potential preferences and offer the most appropriate recommendation based on users, items and relevant information. The recommendation system has a history of decades. With the rapid development of deep learning, recommendation systems based on this technology have also shone in recent years. Deep-learning-based recommendation systems such as DLRM have become classic recommendation system models and have been widely used in the industry.

However, the recommendation system based on deep learning faces some challenges: This model usually has humongous parameters, requiring a large number of sample data and a long training time. Therefore, the deep learning-based recommendation system training is not friendly to companies or institutions with few computing resources. So, these companies may outsource the training process to other institutions with rich computing resources. If the training of the recommendation system is outside the model owner's company, it will raise a problem of data security and privacy protection. Therefore, there is an urgent need for a safe training scheme for recommendation system in-depth learning.

Secure distributed training includes a wide range of concepts, which intend to prevent data leakage to a certain extent in the training process. Federated learning is a new secure distributed deep learning technology [5]. It mainly emphasizes that data will not leave its owner's scope in the training process, and only transfer gradient and other information, ensuring data security at a higher level. Horizontal federated learning is a type of federated learning in which each participant in distributed computing has the same model but different training data.

Security and privacy protection technology is closely related to cryptography technologies and algorithms, such as secure multi-party computing (MPC) [6], homomorphic encryption (HE) [7], differential privacy (DP) [8], trusted execution environment (TEE). MPC requires constructing complex communication schemes with significant communication overhead, while HE needs to consume considerable computing resources to calculate the ciphertext. DP usually affects the model training performance, and it is challenging to balance the impact of privacy protection and training performance. Compared with other privacy protection computing technologies, TEE has the advantages of taking security, compat and performance into account at the same time. TEE can seamlessly support general computing frameworks and applications, with computing performance comparable to plaintext computing. TEE is an essential technology in scenarios involving big data, high-performance and general privacy computing, such as secure and trusted cloud computing, large-scale data confidentiality cooperation and in-depth learning of privacy protection.

Intel introduced SGX2 in Xeon [9], designed to provide user-space TEE with hardware security as a mandatory guarantee, independent of firmware and software security status. In this paper, the term SGX specifically means SGX2. The processor implements a new set of instruction set extensions and memory access control mechanisms to enable isolated operation between different programs and safeguard the user's critical code and data from malware. SGX's TCB includes only hardware, avoiding software-based TCB's software security vulnerabilities and greatly enhancing system security.

The SGX instruction set extensions allocate a portion of the protected area in memory called Enclave Page Cache (EPC). The data in the EPC is encrypted by the Memory Encryption Engine (MEE) inside the CPU. SGX allows applications to create Enclave containers on the EPC to run protected application code segments and ensure that the data therein is isolated from external software, including the operating system. The Enclave can prove its identity to remote authenticators and provide the necessary functionality to secure the key transfer.

## 2   RELATED WORKS

Privacy-Preserving Recommender Systems have also seen many new advances in recent years. For example, Ribero proposed a differential privacy-based approach to implement a federal recommender system [10]. Minto et al. proposed a practical federated recommender system for implicit data under user-level local differential privacy (LDP) [11]. Ben presented secure multi-party protocols that enable several vendors to share their data in a privacy-preserving manner to allow more accurate Collaborative Filtering (CF) [12].

Federated learning is now widely used in industry, and various cryptographic techniques can be used to ensure privacy security in different application scenarios. For example, Google's Gboard uses the federated averaging algorithm to train the data stored in the mobile phone, uses DP to protect the user's privacy, constructs a global model and pushes it back to the mobile phone to predict the next word input by the use [13]. The AI Department of WeBank released the federated learning open-source framework FATE and applied it to financial business [14]. FATE uses HE and MPC to build the underlying security computing protocol. It has currently supported a variety of machine learning algorithms such as logistic regression and deep learning. ByteDance has launched Fedlearner, an open-source framework that uses cryptographic techniques, such as homomorphic encryption, to provide privacy protection to recommender systems like training of Wide & Deep.

## 3   PROBLEM DEFINITION AND CHALLENGES

In this part, we mainly consider two scenarios.

(1) Data providers often have only a portion of the data within the company to build a well-performed recommendation system model. One solution is to train the model among different data providers collaboratively with their own potion of data, as shown in Figure 1 (a). However, this raises the issue of data security.

(2) Some data providers do not have more abundant computing resources. When training recommendation systems, they may need to hand over the training task to cloud service providers with abundant computing resources, as shown in Figure 1 (b). Issues like privacy leakage may arise in the process.

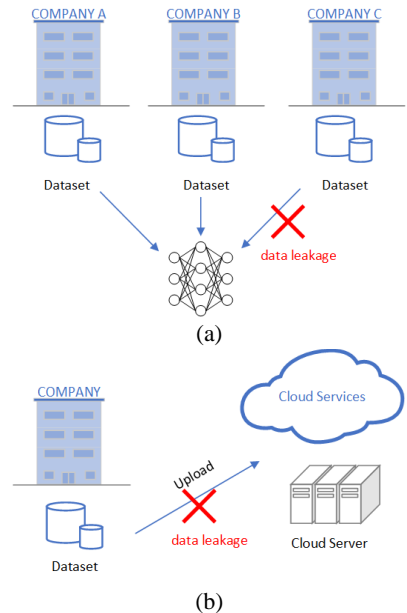

**Figure 1: Two data leakage scenarios**

## 4 METHOD

In this section, we will describe the three aspects of privacy protection we have adopted, as well as two distributed secure training architectures based on DLRM. These two architectures are applied separately to whether data is allowed to leave the local area or not. For the privacy leakage problem in scenario (1) and scenario (2), our Horizontal Federated Learning solution and Secure Distributed Training solution can be used to solve the problem, respectively.

### 4.1 Privacy Protection with Intel SGX

In our solution, privacy protection is provided in the following aspects:

**Runtime security using Intel-SGX.** The gradient information is stored inside the Intel SGX Enclave in the training phase of federated learning. Intel SGX provides assurance that no unauthorized access or memory snooping of the Enclave occurs to prevent leakage of gradient and model information.

**In-Transit security.** We use the Remote Attestation with Transport Layer Security (RA-TLS) of Intel SGX technology to ensure security during transmission [15]. This technology combines TLS technology and remote attestation technology. RA-TLS uses TEE as the hardware root of trust. The certificate and private key are generated in the Enclave and are not stored on the disk. Therefore, participants cannot obtain the certificate and private key in plain text, preventing the man-in-the-middle attacks. In this federated learning solution, RA-TLS is used to ensure the encrypted transmission of gradient information.

**Application integrity and confidentiality.** To solve the problem of how to verify the untrusted application integrity, we implemented RA-TLS enhanced gRPC in TensorFlow to verify the Intel SGX Enclave. It ensures that the runtime application is a trusted version.

### 4.2 DLRM

The following paper will demonstrate our horizontal federated learning and secure distributed training scheme with DLRM. DLRM is a classical neural network-based recommendation model proposed by Facebook for social media and ad click-through rate prediction. The model contains multiple embedding layers and multi-layer-perceptrons that take multiple categorical and numerical data as input and give a probability output.

### 4.3 Horizontal Federated Learning

We proposed a horizontal federated learning solution based on Intel SGX technology for the DLRM. This solution mainly contains the below items:
- AI Framework – TensorFlow, an open-source platform for machine learning.
- Security Isolation LibOS – Gramine, an open-source project for Intel SGX, can run applications with no modification in Intel SGX.
- Platform Integrity - Providing the Remote Attestation mechanism so that users can gain trust in the remote Intel SGX platform.

We adopt the Parameter Server architecture provided by TensorFlow. There are two types of training nodes: parameter server and worker. The parameter server node mainly saves and aggregates the parameter information of the model, and the worker node is mainly responsible for performing forward propagation and backpropagation during the training process.

In the training process, each worker uses local data in its Enclave to complete a round of training and then sends the gradient information in the backpropagation process to the parameter server through the RA-TLS technology. The parameter server completes the gradient aggregation and updates network parameters. Finally, it sends the updated parameters back to each worker.

The training phase can be divided into the following steps as shown in Figure 2:

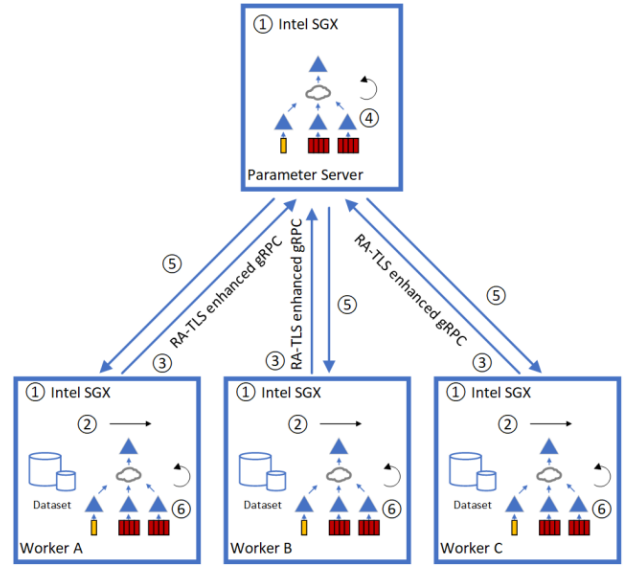

**Figure 2: Architecture of Horizontal Federated Learning based on Intel SGX**

① Using Intel SGX technology, the training program of the participants runs in different Enclaves.

② Workers calculate gradient information based on local data in the Enclave environment.

③ Workers send gradient to parameter server through RA-TLS.

④ Parameter server performs gradient aggregation and updates global model parameters in the Enclave.

⑤ Parameter server sends model parameters to workers.

⑥ Workers update local model parameters.

Steps ②-⑥ will be repeated continuously during the training process. Since the workers and the parameter server run in a memory-encrypted Enclave environment, and RA-TLS technology guarantees encryption during transmission, this solution can ensure privacy during training.

### 4.4 Secure Distributed Training

We implemented another secure distributed training scheme, which divides the distributed nodes into three categories: parameter server, worker, and chief. The chief node has the information of the dataset and model. Before the training starts, the node distributes the information required for training such as model or dataset to other nodes, as shown in Figure 3.

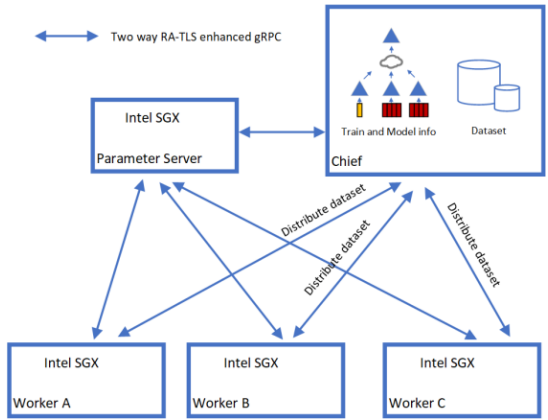

**Figure 3: Architecture of secure distributed training based on Intel SGX**

In terms of privacy protection, we also use RA-TLS enhanced GRPC to ensure communication security and remote authentication during the training process. All three distributed nodes above run in the Enclave environment to ensure runtime security.

The difference between this scheme and the horizontal federated learning is that the data is not stored in the worker node, but the chief node divides the dataset during training initialization and transmits it to each node through the network. This process is also under the RA-TLS enhanced gRPC framework to ensure secure transmission.

## 5 EXPERIMENT

### 5.1 Experiment Configuration

In the experiment, we implement horizontal federated learning and secure distributed training on TensorFlow and test it with the DLRM model. For both tests, we evaluate the training process with click-through record in Kaggle Cretio Ad dataset, which contains 13 numerical features in float32 and 26 categorical features in int32. For both training, we selected one tenth of the original dataset as the training set. The training hyperparameters were set according to the original paper [4]. To justify the correctness and efficiency of our implementation, we collect and compute the loss, training speed and accuracy and compare with non-SGX distributed training.

The model is performed on a dual-socket Intel Ice Lake server with 1TB DDR4 DRAM, which supports Intel SGX. The TensorFlow is optimized with Intel OneAPI for metric acceleration (AVX2, AVX512) [16]. We allocate 32GB of memory for each Enclave and support up to 1024 threads.

### 5.2 Federated Learning Performance

The implementation of horizontal federated learning is based on TensorFlow distributed training. We set up one parameter server and four workers. They each run in a Docker container environment. In this scenario, a worker simulates a company that has its own private dataset.

Figures 4 and 5 are the training losses and accuracy of native distributed training and horizontal federated learning with Intel-SGX. We can see that the converging speed and final accuracy are approximately identical.

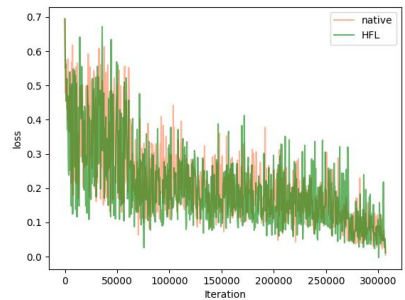

**Figure 4: Training loss of native and HFL**

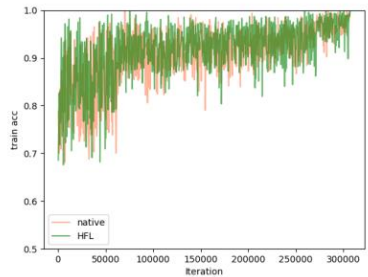

**Figure 5: Training accuracy of native and HFL**

Figure 6 shows the training speed of native distributed training and horizontal federated learning with Intel SGX. Introducing our federated learning scheme brings about a 2.2x overhead.

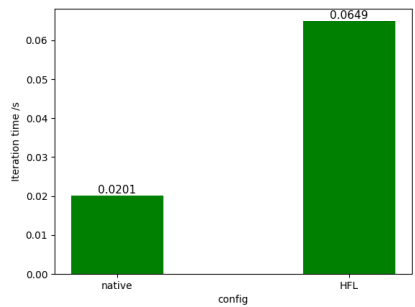

**Figure 6: Training time of native and HFL**

## 5.3 Secure Distributed Training Performance

We set up a chief, a parameter server and two workers to complete the training. They each run in a separate Docker container environment as well. In this scenario, workers and ps can be placed on cloud servers with abundant computing resources, while the chief node can be placed on the data owner. The test is also based on the TensorFlow distributed framework.

Figures 7 is the training losses of native distributed training and secure distributed training with Intel-SGX. We can see that the converging speed and final accuracy are approximately identical.

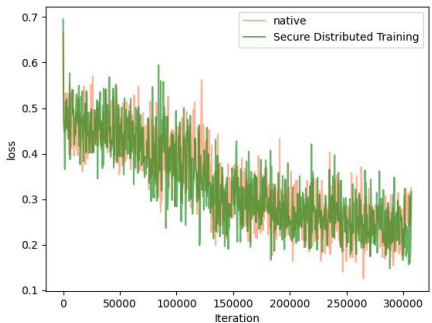

**Figure 7: Training loss of native and Secure Distributed Training**

Figures 8 shows the training speed of native distributed training and Secure Distributed Learning with Intel SGX. Introducing our federated learning scheme brings about a 1.6x overhead.

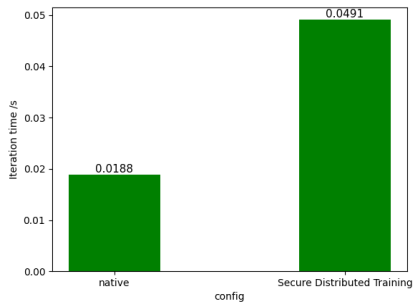

**Figure 8: Training time of native and Secure Distributed Training**

## 5.4 Experiment summary

We can find from the above two experiments based on different architectures and scenarios: adopting our privacy-preserving scheme has almost no effect on the training convergence of DLRM. In addition, the overhead brought by the two schemes is 2.2x and 1.6x, respectively. For MPC, we did not find any available performance data for recommendation system implementations (MPC version vs native version). However, compared with HE-based methods [17], [18] and DP-based methods [19], the overhead of our scheme is tolerable.

## 6 CONCLUSION

This paper introduces security training methods for recommendation systems with Intel-SGX. We raised two scenarios of recommendation model training that face data security challenges. To solve these urgent needs, we preset horizontal federated learning and secure distributed training schemes with Intel-SGX on the TensorFlow. To justify them, we established experiments with the training of the DLRM and evaluated the correctness and performance of our method. The results prove that our implementation is correct in terms of training accuracy and fast compared to other security deep learning schemes.